\documentclass[journal]{IEEEtran}

\ifCLASSINFOpdf
\else
\fi
\usepackage{graphicx}
\usepackage{cite}
\usepackage{amsmath,amssymb,amsfonts,bbding}
\usepackage{algorithmic}
\usepackage{textcomp}
\usepackage[caption=false,font=footnotesize]{subfig}
\usepackage{algorithm}
\usepackage{amsmath,bm}
\usepackage{multirow}
\usepackage{threeparttable}
\usepackage{color}
\usepackage{colortbl}
\usepackage{xcolor}
\usepackage{hyperref}
\usepackage{stfloats}
\usepackage[sort&compress,numbers]{natbib}
\begin{document}

%
\title{RSVG: Exploring Data and Models for Visual Grounding on Remote Sensing Data}

\author{Yang Zhan, Zhitong Xiong,~\IEEEmembership{Member, IEEE}, and~Yuan Yuan,~\IEEEmembership{Senior Member, IEEE}
\thanks{
Yang Zhan and Yuan Yuan are with the School of Artificial Intelligence, Optics, and Electronics (iOPEN), Northwestern Polytechnical University, Xi'an 710072, China (e-mail:\href{mailto:y.yuan@nwpu.edu.cn}{y.yuan@nwpu.edu.cn}).}
\thanks{Zhitong Xiong is with the Chair of Data Science in Earth Observation, Technical University of Munich (TUM), 80333 Munich, Germany.}
}


\maketitle

\begin{abstract}
	In this paper, we introduce the task of visual grounding for remote sensing data (RSVG). RSVG aims to localize the referred objects in remote sensing (RS) images with the guidance of natural language.
	To retrieve rich information from RS imagery using natural language, many research tasks, like RS image visual question answering, RS image captioning, and RS image-text retrieval have been investigated a lot. However, the object-level visual grounding on RS images is still under-explored. Thus, in this work, we propose to construct the dataset and explore deep learning models for the RSVG task. Specifically, our contributions can be summarized as follows. 
	1) We build the new large-scale benchmark dataset of RSVG, termed RSVGD, to fully advance the research of RSVG. This new dataset includes image/expression/box triplets for training and evaluating visual grounding models.
	2) We benchmark extensive state-of-the-art (SOTA) natural image visual grounding methods on the constructed RSVGD dataset, and some insightful analyses are provided based on the results. 
	3) A novel transformer-based Multi-Level Cross-Modal feature learning (MLCM) module is proposed. Remotely-sensed images are usually with large scale variations and cluttered backgrounds. To deal with the scale-variation problem, the MLCM module takes advantage of multi-scale visual features and multi-granularity textual embeddings to learn more discriminative representations. To cope with the cluttered background problem, MLCM adaptively filters irrelevant noise and enhances salient features. In this way, our proposed model can incorporate more effective multi-level and multi-modal features to boost performance. Furthermore, this work also provides useful insights for developing better RSVG models. The dataset and code will be publicly available at \url{https://github.com/ZhanYang-nwpu/RSVG-pytorch}.
	
\end{abstract}


\IEEEpeerreviewmaketitle

\section{Introduction}
\label{sec:introduction}

\IEEEPARstart{W}{ith} the rapid development of remote sensing (RS) technology, the quantity and resolution of RS images have been rapidly improved \cite{earthnets4eo, chen2021remote,9371014}. To efficiently process and retrieve RS imagery, tasks of integrating natural language and RS imagery have become a hot research topic. Although there are many studies combining natural language processing (NLP) with RS, like RS image captioning \cite{8240966, li2021recurrent,zhang2021global}, RS image-text retrieval \cite{mikriukov2022deep,yuan2021lightweight,yuan2021exploring}, and RS image visual question answering \cite{9088993,yuan2022easy,yuan2022change}, the task of visual grounding for RS data (RSVG) is still under-explored.

\begin{figure}[t]	
	\centering
	\includegraphics[width=1\linewidth]{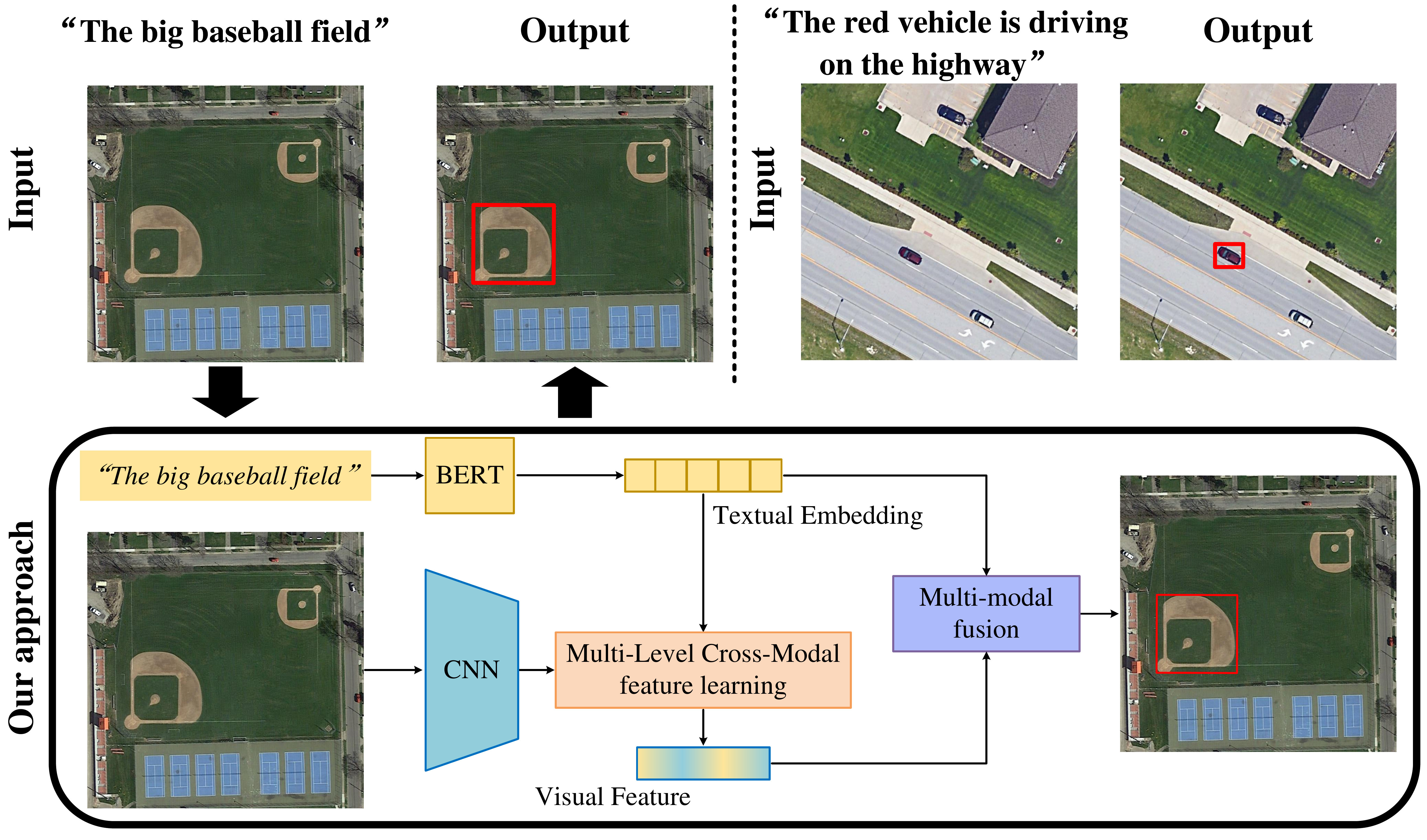}
	\caption{Illustration of our task and approach. Top row: the input is an image-query pair and the output is a bounding box of the referred object. Each pair consists of an RS image and a query expression and the query can be a phrase or a sentence. Bottom row: our approach is an end-to-end transformer-based framework with four steps: 1) multi-modal encoding, 2) multi-level cross-modal feature learning, 3) multi-modal fusion, and 4) localizing.}
	\label{fig:RSVG_task}
\end{figure}

RSVG aims to localize the object referred by the query expression in RS images, as shown in Fig. \ref{fig:RSVG_task}. Given an RS image and a natural language expression, RSVG is asked to provide the referred object's bounding box. Query expressions include phrases and sentences. Multimodal machine learning (MML) \cite{xiong2020variational,xiong2021ask} enables computers to understand image-text pairs. Therefore, RSVG makes it possible for ordinary users, not limited to professionals or researchers, to retrieve objects in RS images, realizing human-computer interaction. It has a wide application prospect in scenarios such as military target detection, military intelligence generation, natural disaster monitoring, agriculture production, search and rescue activities, and urban planning \cite{9088993,8240966,9371014}.

Since RSVG has high potential in real-world applications, this paper explores the novel task and constructs a new large-scale dataset. 
We build a benchmark dataset, named RSVGD, using an automatic generation method with manual assistance. The construction procedure is shown in Fig. \ref{fig:RSVG_dataset}, including four steps: 1) box sampling, 2) attribute extraction, 3) expression generation, and 4) worker verification. The RSVGD dataset is sampled from the target detection dataset DIOR \cite{li2020object}. DIOR is large-scale on the number of object categories, object instances, and images, and has significant object size variations, image quality variations, inter-class similarity, and intra-class diversity. Thus, this new dataset provides researchers with a good data source to foster the research of RSVG. Specifically, RSVGD contains 38,320 RS image-query pairs and 17,402 RS images, and the average length of expressions is 7.47. 
Nowadays, natural image visual grounding has been developed significantly. To fully advance the task of RSVG, we benchmark extensive SOTA visual grounding methods on the RSVGD dataset. The existing methods can be divided into two-stage methods \cite{zitnick2014edge,yu2016modeling,hu2016natural,zhang2018grounding,hu2017modeling,yu2018mattnet,liu2019improving,rohrbach2016grounding,zhuang2018parallel,yu2018rethinking,chen2017query,liu2019learning,hong2022learning,yang2019dynamic,wang2019neighbourhood,yang2020relationship}, one-stage methods \cite{fukui2016multimodal,chen2018real,sadhu2019zero,yang2019fast,liao2020real,yang2020improving,huang2021look,liao2022progressive}, and transformer-based methods \cite{du2021visual,Deng_2021_ICCV,li2021referring,sun2022proposal,yang2022improving, ye2022shifting}. The experimental results show that transferring the visual grounding methods for natural image to RS image can obtain only acceptable results. Even if the above methods have achieved success in the natural domain, they still have some challenges that need to be tackled for the RSVG task.

Based on the characteristics of RS imagery and visual grounding, we propose a Multi-Level Cross-Modal feature learning (MLCM) module, which effectively improves the performance of RSVG.
Firstly, unlike natural scene images, RS images are gathered from an overhead view by satellites, which results in large scale variations and cluttered backgrounds. Due to the characteristics, the model for solving RS tasks has to consider multi-scale inputs. The methods on natural images fail to fully take account of multi-scale features, which leads to suboptimal results on RS imagery. In addition, the background content of RS images contains numerous objects unrelated to the query, but natural images generally have salient objects. Due to the lack of filtering redundant features, the previous models are difficult to understand RS image-expression pairs. Therefore, we attempt to design a network that includes multi-scale fusion and adaptive filtering functions to refine visual features. 
Second, the previous frameworks that extract visual and textual features isolatedly do not conform to human perceptual habits, and such visual features lack the effective information needed for multi-modal reasoning. 
Inspired by the above discussion, we address the problem of how to learn fine-grained semantically salient image representations under multi-scale visual feature inputs. Based on cross-attention mechanism, MLCM module first utilizes multi-scale visual features and multi-granularity textual embeddings to guide the visual feature refining and achieve multi-level cross-modal feature learning. Considering that objects in an RS image are usually correlated, \textit{e.g.}, stadiums usually co-occur with ground track fields, MLCM discovers the relations between object regions based on self-attention mechanism. Specifically, our MLCM includes multi-level cross-modal learning and self-attention learning.
To sum up, our contributions can be summarized in the following aspects:
\begin{enumerate}
	\item To foster the research of RSVG, we design an automatic RS image-query generation method with manual assistance, and the new large-scale dataset is constructed. Specifically, the new dataset contains 38,320 image-query pairs and 17,402 RS images.
	\item We benchmark extensive SOTA natural image visual grounding methods on our RSVGD dataset. Based on experimental results, some analyses about the effects of different methods are given, which provide useful insights on the RSVG task.
	\item To address the problems of scale-variation and cluttered background of RS images and capture the rich contextual dependencies between semantically salient regions, a novel transformer-based MLCM module is devised to learn more transcendent visual representations. MLCM can incorporate effective information from multi-level and multi-modal features, which enables our method to achieve competitive performance.
\end{enumerate}

This paper is organized as follows. We review the related work of natural image visual grounding in Section \ref{sec:relatedwork}. 
In Section \ref{sec:dataset}, the construction procedure of the new dataset is described and the characteristics are analyzed.
In Section \ref{sec:methods}, we present our transformer-based RSVG method.
Evaluation methods and extensive experiment results are shown in Section \ref{sec:experiments}.
Finally, we conclude this work in Section \ref{sec:conclusion}.

\section{Related Work}
\label{sec:relatedwork}
In this section, we comprehensively review the related works about natural image visual grounding methods. To be more specific, two-stage, one-stage, and transformer-based methods are summarized in detail as follows.

\subsection{Two-stage Visual Grounding Methods}
With the development of visual grounding, various two-stage methods have been proposed. \citet{yu2016modeling} introduced better visual context feature extraction methods and found that visual comparison with other objects in the image helps to improve the performance.
In \cite{hu2016natural}, a Spatial Context Recurrent ConvNet (SCRC) is presented, which contains two CNNs to extract local image features and global scene-level contextual features.
\citet{zhang2018grounding} proposed a variational Bayesian method for complex visual context modeling. Besides, a localization score function was also proposed, which is a variational lower bound consisting of multimodal modules of three specific cues and can be trained end-to-end using supervised or unsupervised losses.
\citet{hu2017modeling} attempted to parse the natural language into three modules: subject, relationship, and object, and align these components to candidate regions. The three modules are used to predict the scores of each candidate region. Attention mechanisms have been further introduced \cite{yu2018mattnet,liu2019improving} in each module to better model the interaction between language expressions and candidate regions.
In addition, the attention mechanism \cite{rohrbach2016grounding} is utilized to reconstruct the input phrase and a parallel attention network (ParalAttn) \cite{zhuang2018parallel}, including image-level and proposal-level attention, is proposed.
\citet{yu2018rethinking} found that existing two-stage methods pay more attention to multi-modal representation and region proposals ranking. Therefore, they proposed DDPN to improve region proposal generation, considering both the diversity and discrimination.
\citet{chen2017query} designed a reinforcement learning mechanism to guide the network to select more discriminative candidate boxes.
In addition to the above methods, NMTree \cite{liu2019learning} and RvG-Tree \cite{hong2022learning} utilized tree networks by parsing the language.
To capture object relation information, several researchers \cite{yang2019dynamic,wang2019neighbourhood, yang2020relationship} construct graphs.
\citet{yang2019dynamic} and \citet{wang2019neighbourhood} proposed graph attention network to accomplish visual grounding. 
CMRIN \cite{yang2020relationship} utilized Gated Graph Convolutional Network to fuse multimodal information.

\subsection{One-stage Visual Grounding Methods}
One-stage methods are more computation-efficient and can avoid error accumulation in multi-stage frameworks. Thus, many one-stage methods have been investigated.
Some works use CNN and LSTM or Bi-LSTM to extract visual features and textual features \cite{fukui2016multimodal,chen2018real,sadhu2019zero}.
Multimodal Compact Bilinear pooling (MCB) is first proposed in \cite{fukui2016multimodal} to fuse the multi-modal features.
\citet{chen2018real} designed a multimodal interactor to summarize the complex relationship between visual features and textual features. Besides, a new guided attention mechanism was designed to focus visual attention on the central area of the referred object.
In \cite{sadhu2019zero}, multi-scale features are extracted and multi-modal features are fed to the fully convolutional network to regress box coordinates.
Significant improvement is observed as \citet{yang2019fast} fused textual embeddings with YOLOv3 detector results and augmented the visual features with spatial features.
\citet{liao2020real} defined the visual grounding problem as a correlation filtering process. They mapped textual features into three filtering kernels and performed correlation filtering on the image feature map.
To address the limitations of FAOA \cite{yang2019fast} in complex queries for visual grounding, \citet{yang2020improving} proposed a recursive sub-query construction (ReSC) network.
The latest one-stage methods \cite{huang2021look,liao2022progressive} focus on visual branching and use language expression to guide the visual feature extraction.
A landmark feature convolution module \cite{huang2021look} is designed to transmit visual features under the guidance of language and encode spatial relations between the object and its context.
\citet{liao2022progressive} proposed a language-guided visual feature learning mechanism to customize visual features in each stage and transfer them to the next stage.

\subsection{Transformer-based Visual Grounding Methods}
Recently, transformer-based methods have attracted more and more research attention due to the high efficiency and visual grounding performance.
\citet{du2021visual} and \citet{Deng_2021_ICCV} proposed the earliest end-to-end transformer-based visual grounding network, \textit{i.e}, VGTR and TransVG.
VGTR \cite{du2021visual} was a transformer structure that can learn visual features under the guidance of expression. TransVG \cite{Deng_2021_ICCV} was a network stacked with multiple transformers, including BERT, visual transformer, and multimodal fusion transformer.
Some studies \cite{li2021referring,sun2022proposal} propose a multi-task framework. \citet{li2021referring} utilized transformer encoder to refine visual and textual features and designed a query encoder and decoder for referring expression comprehension (REC) and segmentation (RES) at the same time.
\citet{sun2022proposal} proposed the transformer model for REC and referring expression generation (REG), which uses the same cross-attention module and fusion module to perform multi-modal interaction.
Similar to the latest one-stage methods, the latest transformer-based methods \cite{yang2022improving,ye2022shifting} also focus on the improvement of visual branches and adjusting visual features by combining multi-modal features. 
VLTVG \cite{yang2022improving} aims to adjust visual features with a visual-linguistic verification module and aggregate visual context with a language-guided context encoder. The core of these modules is multi-head attention. QRNet \cite{ye2022shifting} contains a language query aware dynamic attention mechanism and a language query aware multi-scale fusion to adjust visual features.

\section{Dataset Construction}
\label{sec:dataset}
In this section, we will introduce the construction procedure of the new dataset in Section \ref{sec:RSVGdataset}. The statistical analysis of our RSVGD is shown in Section \ref{sec:dataanalysis}.

\subsection{RSVGD: a new dataset for RSVG}
\label{sec:RSVGdataset}
\begin{figure}[t]	
		\centering
		\includegraphics[width=0.95\linewidth]{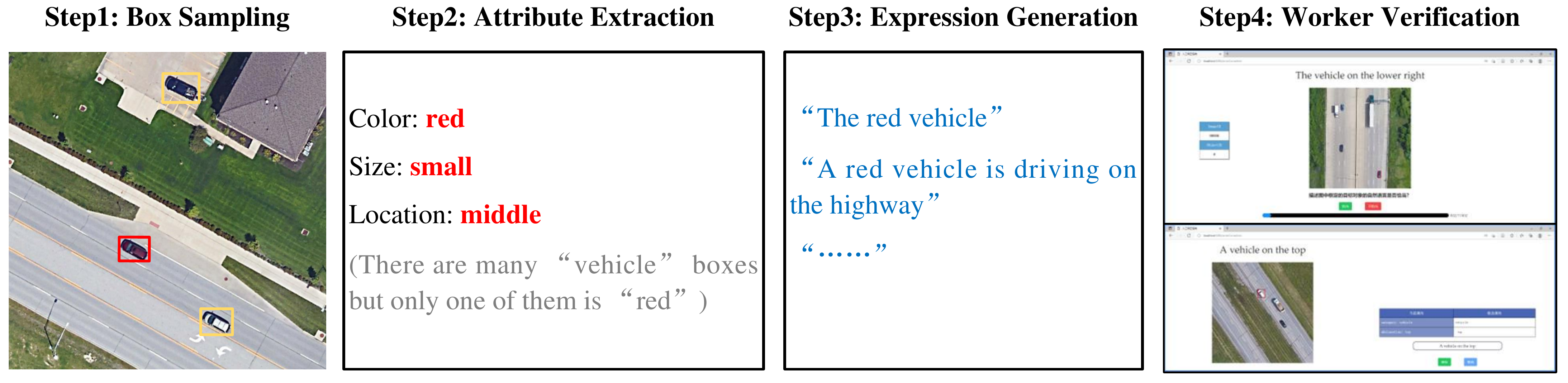}
		\caption{Illustration of the dataset construction processes. Step 1: the red box is the sampling result; yellow boxes are ignored. Step 2: attribute extraction examples in the previous RS image. Step 3: expression generation examples in the previous RS image. Step 4: the dataset is manually validated using a data correction system.}
		\label{fig:RSVG_dataset}
	\end{figure}

The dataset for RSVG requires lots of RS images with the annotation and description of different objects. Therefore, we utilize the existing target detection dataset DIOR \cite{li2020object} as the basic data to construct a new benchmark dataset.
Over the years, various visual grounding datasets \cite{young2014image, guadarrama2014open, kazemzadeh2014referitgame, plummer2015flickr30k, yu2016modeling,krishna2017visual,de2017guesswhat, johnson2017clevr, nguyen2019object, liu2019clevr, deruyttere2019talk2car, wang2020give, yang2020graph, chen2020cops, wu2020phrasecut, liu2021refer} based on real-world and computer-generated images have been proposed to study visual grounding. The construction methods of each dataset are divided into manual annotation \cite{young2014image,guadarrama2014open,plummer2015flickr30k, krishna2017visual, deruyttere2019talk2car,wang2020give,liu2021refer}, game collection \cite{kazemzadeh2014referitgame, yu2016modeling, de2017guesswhat}, and automatic generation \cite{johnson2017clevr, liu2019clevr,yang2020graph, chen2020cops,wu2020phrasecut}. We design an automatic image-query generation method with manual assistance to collect image/expression/box triplets, as shown in Fig. \ref{fig:RSVG_dataset}. A detailed description of the generation of different query expressions is given in what follows.

\textbf{Step 1: Box sampling.}
DIOR dataset includes 23,463 RS images, 192,472 object instances, and 20 object categories. The image size is 800 $\times$ 800 pixels and the spatial resolution range is from 0.5m to 30m. 
First, the data containing annotation errors in the DIOR dataset are removed, \textit{e.g.} axis-aligned bounding box coordinates $x_{min}  \ge x_{max}$ or $y_{min}  \ge y_{max}$. $\left ( x_{min}, y_{min}, x_{max}, y_{max}  \right ) $ is the coordinate of the ground-truth bounding box. Then, bounding boxes that are less than 0.02\% or greater than 99\% of the image size are also removed. Finally, we sample no more than 5 objects of the same category in each RS image to avoid unclear references of expression caused by many of the same category of objects in the image.

\textbf{Step 2: Attribute extraction.}
By analyzing visual grounding datasets from the real world, such as RefItGame \cite{kazemzadeh2014referitgame}, RefCOCO \cite{yu2016modeling}, and RefCOCO+ \cite{yu2016modeling}, a set of attributes widely contained in referring expressions is summarized. we extract the attribute set and define it as a 7-tuple $A=\left \{ a_{1},a_{2},a_{3},a_{4},a_{5},a_{6},a_{7} \right \}$. The symbol, type, and example of each attribute are shown in Table \ref{tab:attribute_info}. 
The object category can be obtained directly from the DIOR dataset. The HSV color recognition method is used to obtain the object's color. Object size is measured by the ratio of bounding box area to image size. The geometry attribute is set in advance for some objects with a fixed shape, such as rectangular basketball courts, circular storage tanks, etc. The geometric attribute of some objects that do not have a describable geometry is empty, such as airports, golf fields, etc. For other objects, we use relevant functions in the OpenCV library to extract object contours for common geometry recognition. Besides, the length and width of the bounding box are also combined to judge whether the object is slender or square. The absolute location refers to the location of the object in the image, which can be judged by the coordinates of bounding boxes. The above attributes $\left \{ a_{1},a_{2},a_{3},a_{4},a_{5} \right \}$ belong to the object's own attributes and the relationship attributes are $\left \{ a_{6},a_{7} \right \}$. The relative location and relative size relation allow expressions to be associated with another object. The relative location relation is obtained by comparing the coordinates of bounding boxes and center points. The relative size relation is determined by comparing two objects’ ratios of the bounding box area to the image size.

	\begin{table}[]
		\centering
		\caption{Specific information for each attribute.}
		\begin{tabular}{lll}
			\hline
			& Attribute                  & Example                                    \\ \hline
			$a_{1}$ & category                   & (\textit{e.g.} “plane, ship”)                        \\
			$a_{2}$ & color                      & (\textit{e.g.} “blue, red”)                          \\
			$a_{3}$ & size                       & (\textit{e.g.} “tiny, big”)                          \\
			$a_{4}$ & geometry                  & (\textit{e.g.} “square, round”)                      \\
			$a_{5}$ & absolute location          & (\textit{e.g.} “top of the image”)                   \\
			$a_{6}$ & relative location relation & (\textit{e.g.} “the car is on the left of the tree”) \\
			$a_{7}$ & relative size relation     & (\textit{e.g.} “the car is smaller than the tree”)   \\ \hline
		\end{tabular}\label{tab:attribute_info}
	\end{table}

\textbf{Step 3: Expression generation.}
To make the generated query expressions representative of the natural language used in the real world, we pre-set textual templates following the Cops-Ref \cite{chen2020cops} dataset. The filling of the textual template is the expression generation process. 
Textual templates include the phrase template and the sentence template. The phrase template uses the object's own attributes $\left \{ a_{1},a_{2},a_{3},a_{4},a_{5} \right \}$ in the following form:

$The/A$ $\left ( att_{0} \right ) $ $\left ( obj_{0} \right ) $  $\Rightarrow  $ $The/A$ $\langle a_{2}\rangle$ $\langle a_{3}\rangle$ $\langle a_{4}\rangle$ $a_{1}$ $\langle in/on$ $the$ $a_{5} \rangle$ .
	
The left of $\Rightarrow $ is the phrase template and the right is the example filled in with specific attributes. The attribute that can be null is bounded with $\left \langle  \right \rangle $ and attributes $\left \{ a_{2},a_{3},a_{4} \right \}$ can be filled in any order. The sentence template uses the relationship attributes $\left \{ a_{6},a_{7} \right \}$ to relate two objects in the following form:

$The/A$ $\left ( att_{0} \right ) $ $\left ( obj_{0} \right ) $ $is$ $a_{6}/a_{7}$ $the$ $\left ( att_{1} \right ) $ $ \left ( obj_{1} \right ) $ .

We select textual templates and fill attributes to generate a query expression for each bounding box. The generation algorithm may be summarized as the following few steps:
	\begin{enumerate}
		\item We first check if the selected object category is unique in the RS image. If so, we fill the phrase template with the category name and randomly selected object attributes.
		\item If the object category is not unique, we look for unique attributes of the object to distinguish it from other objects of the same category. If such an object attribute exists, we combine it with the category name to fill the phrase template.
		\item If no such unique object attribute exists, we look for distinguishable relationship attributes. If such a relationship attribute exists, we combine it with the attributes of two objects to fill the sentence template.
		\item If all the above fail, the object is discarded.
	\end{enumerate}
	
\textbf{Step 4: Worker verification.}
Due to the complex backgrounds and numerous objects of RS images, attribute extraction may be wrong, especially for color and geometry. In addition, we use box regions instead of object pixel regions, which may cause errors in size attribute and relative size relation. Coupled with the unreliability of simple judgments of absolute location and relative location relation, the expression may be ambiguous. Therefore, RSVGD requires worker verification to help correct errors or ambiguous language expressions. The worker verification method consists of two main strategies, majority voting \cite{snow2008cheap} and rapid judgments \cite{krishna2016embracing}. We only use rapid judgments to speed up the validation of datasets. To improve the efficiency, we develop a dataset manual correction system, and the system interface is shown in Fig. \ref{fig:RSVG_dataset}.

\subsection{Data analysis}
\label{sec:dataanalysis}
We construct a large-scale RSVGD, where each object instance in the RS image corresponds to a unique language expression. Our constructed RSVGD consists of 38,320 language expressions across 17,402 RS images and contains 20 object categories. The average length of expressions is 7.47 and the size of the vocabulary is 100. We now present a more detailed statistical analysis of the RSVGD dataset.

\begin{figure*}[t]	
		\centering
		\includegraphics[width=0.9\linewidth]{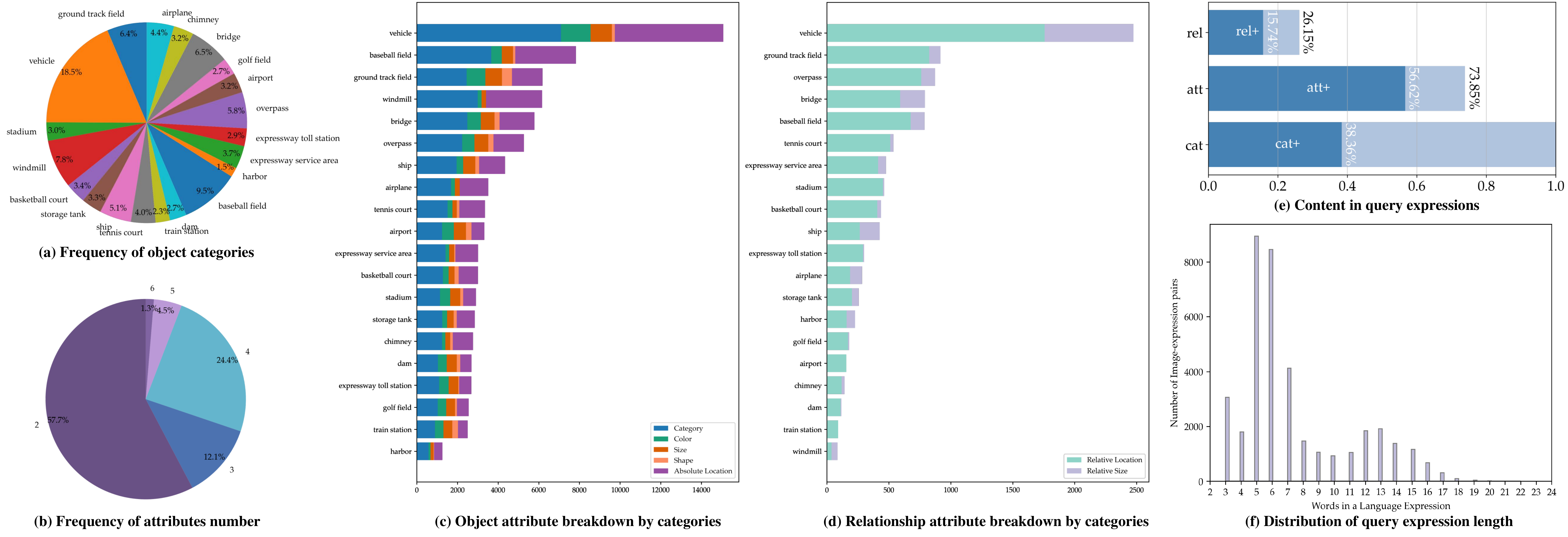}
		\caption{Statistical analysis of the constructed RSVGD. (a) shows the frequency of each object category in RSVGD. (b) shows the frequency of the number of attributes contained in each query expression. (c) shows the frequency of the object attribute occurrence for each category. (d) shows the frequency of the relationship attribute occurrence for each category. (e) shows the distribution of expressions that have category (cat) and expressions that can distinguish objects by category alone (cat+), and similarly for attributes and relationships. (f) shows the distribution of the length of query expressions.}
		\label{fig:analysis}
	\end{figure*}

\begin{figure}[t]	
	\centering
	\includegraphics[width=1\linewidth]{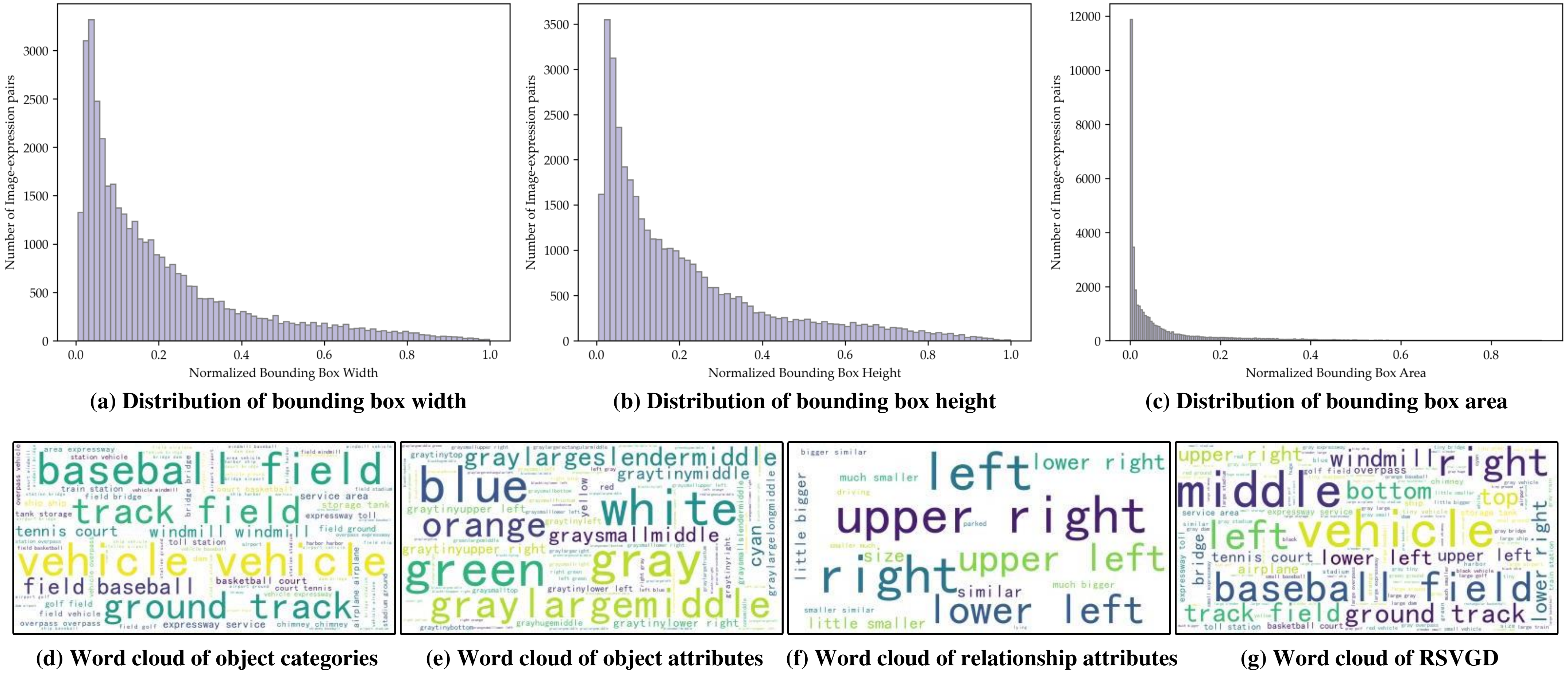}
	\caption{Statistical analysis of the proposed RSVGD. Top row: the distribution of bounding box width (a), bounding box height (b), and bounding box area (c) in the RSVGD. Bottom row: word clouds of categories (d), attributes (e), relationships (f), and the RSVGD (g). The size of each word is proportional to its frequency in the dataset.}
	\label{fig:analysis2}
	\end{figure}

Fig. \ref{fig:analysis} (a) shows the proportion of the number of each object category. The vehicle and harbor are respectively the most and least in the dataset, and the remaining 18 categories account for a relatively uniform proportion, all between 2\%-10\%.
Fig. \ref{fig:analysis} (b) provides the proportion of the number of attributes that appear in each expression. We find that most expressions used two attributes, followed by four attributes, with very few expressions containing five or six attributes.
Fig. \ref{fig:analysis} (c) and (d) respectively show the proportion of object attributes and relationship attributes in query expressions of each object category. The percentage of each attribute is similar in different object categories, so the use of different attributes doesn't depend on the object category.
The bar chart shown in Fig. \ref{fig:analysis} (e) shows three kinds of information about query expressions from bottom to top: the proportion of expressions having object category information (cat) and the proportion of expressions that can distinguish objects by category information alone (cat+), and similarly for attributes and relationships. Specifically, 38.36\% of objects can be distinguished by the object category alone (cat+), 56.62\% of objects can be distinguished by the object's own attribute(att+), and 15.74\% by a relationship attribute (rel+).
Fig. \ref{fig:analysis} (f) shows the distribution of the length of query expressions. The average length of expressions is 7.47 words, with a minimum of 3 words and a maximum of 22 words.
The expressions need to be specific enough to describe individual objects in the RS image, such as the query ``a dam'', but they also need to be general enough to describe high-level concepts in the RS image. Specifically, covering most of the areas in the RS image is often a general description of the image, while covering only a small part of the image is often more specific. The top row of Fig. \ref{fig:analysis2} shows the distribution of the width, height, and area of the bounding box, with the area mainly within 20\% of the RS image.
The bottom row of Fig. \ref{fig:analysis2} shows the word clouds of object categories, object attributes, relationship attributes, and the RSVGD dataset. We can see that RSVGD covers a wide range of objects, with the vehicle, baseball field, and ground track field being the most common object names. The most common object attributes are color (\textit{e.g.}, blue, white, green, and gray), size (\textit{e.g.}, large), and absolute position (\textit{e.g.}, middle), and the most common relationship attributes are relative location relation (\textit{e.g.}, upper right and left).

\section{Methods}
\label{sec:methods}
This section introduces the transformer-based RSVG framework and our proposed MLCM module. We first overview the overall framework in Section \ref{sec:overview}. Then, we elaborate the architecture of the framework and our designs of MLCM module in Section \ref{sec:MLCMF}. Finally, Section \ref{sec:loss_fun} details the loss function of our framework for training.
	
	\begin{figure*}[t]	
		\centering
		\includegraphics[width=0.8\linewidth]{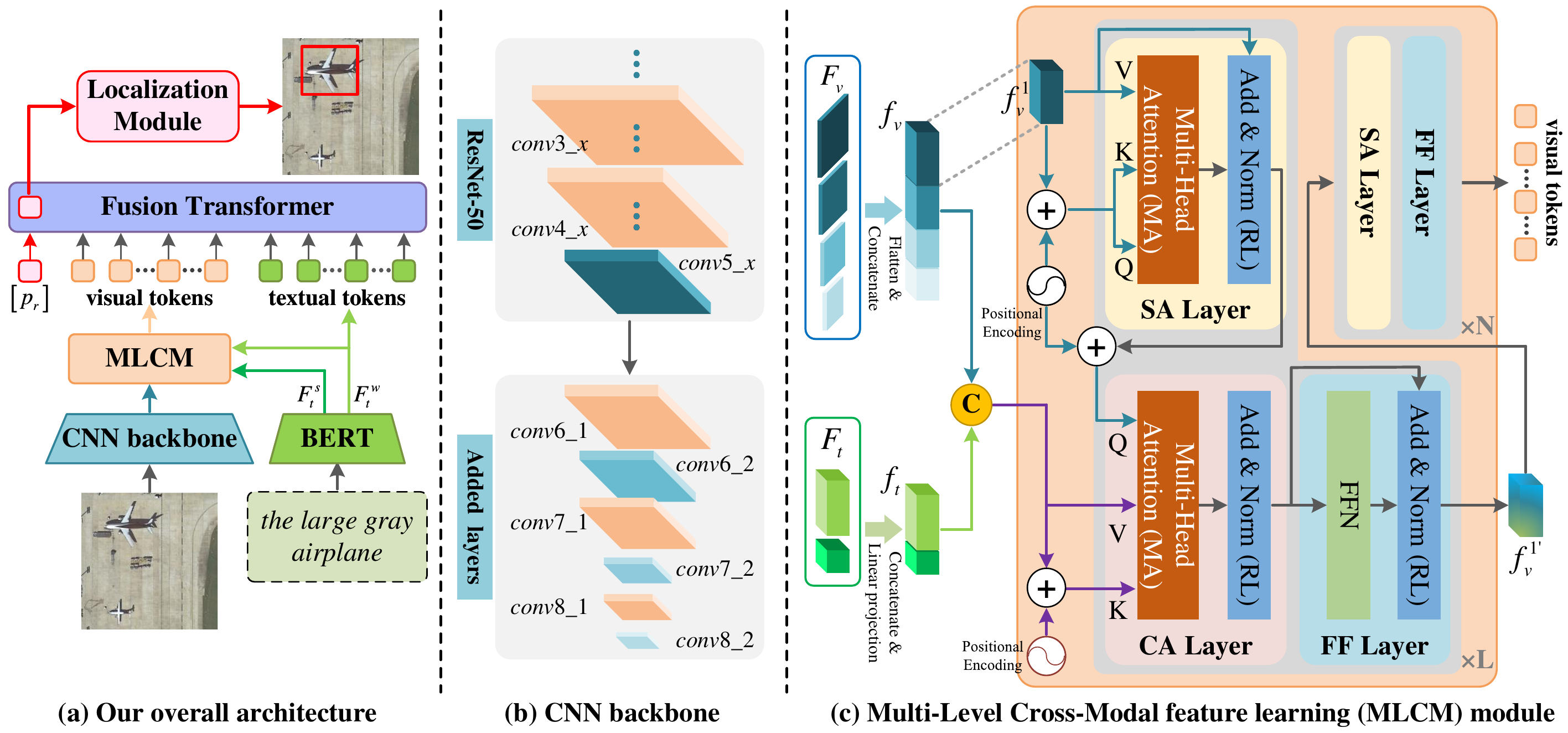}
		\caption{(a) The architecture of the transformer-based RSVG framework. The framework consists of four components: a multimodal encoder, an MLCM module, a multimodal fusion module, and a localization module. (b) Illustration of our CNN backbone. It contains a truncated ResNet-50 and 6 additional convolution layers. (c) Illustration of our proposed MLCM module. It includes multi-level cross-modal learning and self-attention learning. MLCM first utilizes multi-scale visual features and multi-granularity textual embeddings to adaptively filter irrelevant noise and learn more discriminative visual representations. Then, MLCM captures the rich contextual dependencies between semantically salient regions based on self-attention mechanism.
		}
		\label{fig:MLCMF}
	\end{figure*}

\subsection{Overview}
\label{sec:overview}
RSVG is a task about localizing a target object described by a natural language expression in RS images. Our goal is to deal with the problems of scale-variation and the cluttered background of RS images and capture the rich contextual dependencies between semantically salient regions. To this end, we propose an MLCM module to adaptively filter irrelevant noise and discover the relations between object regions, so that visual representations are focused on the valid regions referred by query expressions. 
To get full clues from multi-modal features at different semantic levels, MLCM takes advantage of multi-scale visual features and multi-granularity textual embeddings to refine visual features. 
We design CNN backbone to extract multi-scale visual features and use BERT to obtain word-level and sentence-level textual embeddings. 
Multi-scale visual features contain coarse-scale semantic information and fine-scale detailed information. Multi-granularity textual embeddings contain local and global information from different aspects.
To mine the potential relationship between text semantics and visual perception, we use the transformer-based Multimodal Fusion Module that is analogous to TransVG \cite{Deng_2021_ICCV} model.
Since the visual grounding inference requires more detailed information, we take the refined visual features and word-level textual embeddings as the input of Multimodal Fusion Module. Two linear projection layers are applied to map visual features and textual embeddings into the same dimension. A learnable embedding (a learnable token) is pre-appended to visual embeddings and textual embeddings. It gathers the intra-modal and inter-modal information through Transformer's self-attention mechanism to facilitate visual grounding. 
Finally, the learnable token is sent to Localization Module for the regression of box coordinates. The overall framework is shown in Fig. \ref{fig:MLCMF} (a). We will introduce each module as follows in detail.

\subsection{Multi-level Cross-modal Fusion}
\label{sec:MLCMF}
First, we denote the cross-modal RS image-query dataset as $\mathcal{O} =\left \{ \left ( i_{m} , s_{m}  \right )  \right \} _{m=1}^{M} $ that has \textit{M} image-query pairs. RS images $\mathcal{I}=\left \{  i_{m}\right \} _{m=1}^{M} $ and queries $\mathcal{S}=\left \{  s_{m}\right \} _{m=1}^{M} $ have \textit{M} instances in each modality. To simplify the notations, we denote $I$ and $S$ as single instances of image and text modality, respectively.
	
\textbf{Multimodal Encoder.}
Given an RS image $\boldsymbol{I}\in \mathbb{R}^{H_{0}\times W_{0}\times 3}$ and a query expression $S=\left \{ w_{n} \right \}_{n=1}^{N}$ (\textit{N} is the sentence length) as input  of multimodal encoder, where ${H_{0}\times W_{0}\times 3}$ denotes the size of the RS image and $w_{n}$ represents the $n$-th word.
An overview of our CNN backbone is shown in Fig. \ref{fig:MLCMF} (b). We forward the RS image into the ResNet-50 that removed the average pool and FC layer to generate a 2D visual feature map $\boldsymbol{F}_{v}^{1} \in \mathbb{R}^{H\times W\times C}$. Multi-scale visual features are extracted by adding six additional convolution layers ($conv6\_1$, $conv6\_2$, $conv7\_1$, $conv7\_2$, $conv8\_1$, and $conv8\_2$) to the end of the truncated ResNet-50. $conv6\_1$, $conv7\_1$, and $conv8\_1$ are all designed with 128 filters with a 1$\times$1 filter size and a stride of 1, whereas $conv6\_2$, $conv7\_2$, and $conv8\_2$ are all designed with 256 filters with a 3$\times$3 filter size and a stride of 2. 
$conv6\_2$, $conv7\_2$, and $conv8\_2$ output feature maps of size 9$\times$9$\times$256, 4$\times$4$\times$256, 1$\times$1$\times$256 respectively, which are denoted by $\left [ \boldsymbol{F}_{v}^{2}, \boldsymbol{F}_{v}^{3}, \boldsymbol{F}_{v}^{4} \right ]$. $\boldsymbol{F}_{v}^{1}$ is also transformed into the same channel dimension $c$ = 256, and $F_{v}$ represents the multi-scale visual features:
	
\begin{equation}\label{eq:multi-scale}
	F_{v}=\left [ \boldsymbol{F}_{v}^{1}, \boldsymbol{F}_{v}^{2}, \boldsymbol{F}_{v}^{3}, \boldsymbol{F}_{v}^{4} \right ].
\end{equation}
	
For the expression, we first embed each word $w_{n}$ into a one-hot embedding vector. Then, we convert each one-hot vector into a language token and append $\left [ CLS \right ]$ token and $\left [ SEP \right ]$ token following the common approach in \cite{vaswani2017attention,dehghani2018universal,kenton2019bert,Deng_2021_ICCV}. To capture local semantic information and global sentence contextual information, we use the pre-trained BERT model \cite{kenton2019bert} to extract word-level textual embeddings and sentence-level textual embeddings. BERT contains 12 Transformer encoders and the output channel dimension of BERT is $b$. Specifically, the average of the last four layers' hidden states is taken as the word-level textual embedding $\boldsymbol{F}_{t}^{w} \in \mathbb{R}^{N_{t} \times b}$ of this 
query. Here,  $N_{t}$ represents the length of language tokens and the tokens ensure a fixed length $N_{t}$ by padding or cutting. The embeddings output by the BERT are pooled as the sentence-level textual embedding $\boldsymbol{F}_{t}^{s} \in \mathbb{R}^{1 \times b}$. $F_{t}$ represents the multi-granularity textual embeddings:
\begin{equation}\label{eq:multi-grained}
	F_{t}=\left [ \boldsymbol{F}_{t}^{w}, \boldsymbol{F}_{t}^{s} \right ].
\end{equation}

\textbf{MLCM Module.}
Unlike the traditional Transformer encoder-decoder structure, our MLCM first has a separate decoder, which is connected to a separate encoder. As shown in Fig. \ref{fig:MLCMF} (c), our MLCM consists of two parts: a multi-level cross-modal layer and a self-attention layer.
MLCM requires two inputs, $x$ and $y$. It can refine $x$ by selecting and aggregating valid information from $y$ based on the global relationship between $x$ and $y$ at all positions. In order to refine $x$ under the guidance of multi-level and multi-modal features, the input $y$ should contain information for all levels and modalities. 
We flatten multi-scale visual features $F_{v}$ into $[\boldsymbol{f}_{v}^{1}, \boldsymbol{f}_{v}^{2},\boldsymbol{f}_{v}^{3}, \boldsymbol{f}_{v}^{4}]$, where $\boldsymbol{f}_{v}^{i}\in \mathbb{R}^{N_{i} \times c}$, $N_{i}=H_{i}\times W_{i}$, and $i\in \left \{ 1,2,3,4 \right \} $. Although $N_{i}$ is different, the channel dimension is the same. Instead of sampling to the same size for feature fusion, we directly concatenate $\boldsymbol{f}_{v}^{1}$, $\boldsymbol{f}_{v}^{2}$, $\boldsymbol{f}_{v}^{3}$, and $\boldsymbol{f}_{v}^{4}$ on the channel dimension to obtain $\boldsymbol{f}_{v}\in \mathbb{R}^{N_{1234} \times c}$, where $ N_{1234}={\textstyle \sum_{i=1}^{4}} N_{i} $. The object scale of RS imagery varies greatly. The method maintains the resolution of original features and can preserve useful information about objects of different scales. Word-level features play an important role in visual grounding, especially the word-level feature directly related to the target object in the text. In addition, contextual semantic information at the sentence level can also provide useful clues for visual grounding. Similarly, we concatenate $ \boldsymbol{F}_{t}^{w}$ and $\boldsymbol{F}_{t}^{s}$, and use a linear layer to get $\boldsymbol{f}_{t}\in \mathbb{R}^{\left( N_{t}+1\right) \times c}$. After intra-modal concatenation, we concatenate inter-modal features $\boldsymbol{f}_{v}$ and $\boldsymbol{f}_{t}$ to obtain $\boldsymbol{f}_{vt} \in \mathbb{R}^{\left(N_{1234}+ N_{t}+1\right) \times c}$. As the input $y$, $\boldsymbol{f}_{vt}$ provides rich information on all levels and modalities. We express the L-layer decoder’s process of $l$-th layer as
	\begin{equation}\label{eq:VFRM}
		\boldsymbol{f}_{v}^{1l} = DE^{l} \left ( \boldsymbol{f}_{v}^{1\left ( l-1 \right ) },  \boldsymbol{f}^{vt} \right ), 
	\end{equation}
where $l\in \left [ 1,\dots ,L \right ] $, $\boldsymbol{f}_{v}^{1l}$ is the output of  $l$-th layer. $\boldsymbol{f}_{v}^{1} \in \mathbb{R}^{N_{1} \times c}$ is the initial visual feature as input $x$ of the 1-th layer, where $N_{1} = H \times W$. 
The decoder includes self-attention (SA) layer which refines itself, cross-attention (CA) layer which aggregates complementary information in $\boldsymbol{f}_{vt}$, and feed forward (FF) layer. The specific formula of $ DE^{l}\left(\cdot  \right) $ is as follows:
	\begin{equation}\label{eq:SA}
		\boldsymbol{F}_{sa} =SA\left ( \boldsymbol{f}_{v}^{1\left ( l-1 \right ) }  \right ), 
	\end{equation}
	\begin{equation}\label{eq:CA}
		\boldsymbol{F}_{ca} =CA\left ( \boldsymbol{F}_{sa}, \boldsymbol{f}_{vt}  \right ),
	\end{equation}
	\begin{equation}\label{eq:FF}
		\boldsymbol{f}_{v}^{1l} =FF\left ( \boldsymbol{F}_{ca}  \right ). 
	\end{equation}

The SA layer and CA layer both contain a multi-head attention (MA) module and a residual connection and layer normalization (RL) block. In the MA module, attention is calculated $h$ times. The single attention takes Query $\boldsymbol{Q}$, Key $\boldsymbol{K}$, and Value $\boldsymbol{V}$ as input and is calculated by: 
	\begin{equation}\label{eq:MA}
		Att\left (\boldsymbol{Q},\boldsymbol{K},\boldsymbol{V} \right ) =softmax\left ( \frac{\boldsymbol{Q}\boldsymbol{K}^{T} }{\sqrt{d_{K} } }  \right ) \boldsymbol{V},
	\end{equation}
where $\boldsymbol{Q} \in \mathbb{R}^{N_{Q} \times d}$, $\boldsymbol{K} \in \mathbb{R}^{N_{M} \times d}$, $\boldsymbol{V} \in \mathbb{R}^{N_{M} \times d}$. $N_{Q}$ is the length of $\boldsymbol{Q}$ and $N_{M}$ is the length of $\boldsymbol{K}$ and $\boldsymbol{V}$. $d_{K}$ is the dimension of $\boldsymbol{K}$. The output of $ Att\left(\cdot  \right) $ is the same size $\mathbb{R}^{N_{Q} \times d}$ as $\boldsymbol{Q}$. For each attention, $\boldsymbol{Q}$ and $\boldsymbol{K}$ are appended with their corresponding positional encodings.
	
	\begin{equation}\label{eq:SA2}
		\begin{split}
		SA\left ( \boldsymbol{x} \right ) 
		& =RL\left ( Att\left ( \tilde{\boldsymbol{x}} ,\tilde{\boldsymbol{x}} ,\boldsymbol{x} \right )  \right )  \\
		& = norm\left ( \boldsymbol{x} + Att\left ( \tilde{\boldsymbol{x}} ,\tilde{\boldsymbol{x}} ,\boldsymbol{x} \right )  \right ) = \boldsymbol{x}_{sa},
		\end{split}
	\end{equation}

	\begin{equation}\label{eq:CA2}
		\begin{split}
		CA\left ( \boldsymbol{x}_{sa}, \boldsymbol{y} \right ) 
		& =RL\left ( Att\left ( \tilde{\boldsymbol{x}_{sa}} ,\tilde{\boldsymbol{y}} ,\boldsymbol{y} \right )  \right )   \\
		& =norm\left (\boldsymbol{x}_{sa} + Att\left ( \tilde{\boldsymbol{x}_{sa}} ,\tilde{\boldsymbol{y}} ,\boldsymbol{y} \right )  \right ) =\boldsymbol{x}_{ca},
		\end{split}
	\end{equation}
where $\tilde{\boldsymbol{x}}$, $\tilde{\boldsymbol{x}_{sa}}$, and $\tilde{\boldsymbol{y}}$ are $\boldsymbol{x}$, $\boldsymbol{x}_{sa}$, and $\boldsymbol{y}$ with positional encoding, respectively.

	\begin{equation}\label{eq:x1}
		\tilde{\boldsymbol{x}} = \boldsymbol{x} + PosEncoding\left( \boldsymbol{x}\right), 
	\end{equation}
	
	\begin{equation}\label{eq:x2}
		\tilde{\boldsymbol{x}_{sa}} = \boldsymbol{x}_{sa} + PosEncoding\left( \boldsymbol{x}\right), 
	\end{equation}
	
	\begin{equation}\label{eq:y}
		\tilde{\boldsymbol{y}} = \boldsymbol{y} + PosEncoding\left( \boldsymbol{y}\right), 
	\end{equation}
where $PosEncoding\left(\cdot  \right) $ denotes the function to get positional encoding. The positional encoding of $\boldsymbol{f}_{vt}$ is obtained by concatenating positional encoding of multi-level multi-modal features in sequence. 
	 
The FF layer contains a feed forward network (FFN), which consists of two linear layers and a ReLU activation function in the middle, and an RL block. FF is defined as below: 
	
	\begin{equation}\label{eq:FF2}
	    \begin{split}
		FF\left ( \boldsymbol{x_{ca}} \right )  
		&=RL\left ( FFN\left ( \boldsymbol{x_{ca}} \right )  \right ) \\ 
		&=norm\left (\boldsymbol{x_{ca}} + FFN\left (\boldsymbol{x_{ca}}\right )  \right ).
		\end{split}
	\end{equation}

Considering that objects in an RS image are usually correlated, MLCM discovers the relations between object regions based on self-attention mechanism. Specifically, we build an N-layer encoder. The encoder consists of 6 transformer encoder layers, including 8 MA layers. The output channel sizes of the two fully connected layers in FFN are 2048 and 256 respectively. Through self-attention layers, the output $\boldsymbol{f}_{v}^{1'}$ of the L-layer decoder can see other information in the same feature map. Meanwhile, self-attention layers generate visual embedding for the multimodal fusion module.

\textbf{Multimodal Fusion Module.}
The visual tokens generated by MLCM and the word-level textual embeddings $\boldsymbol{f}_{t}^{w} \in \mathbb{R}^{ N_{t} \times c}$ serve as the input of the fusion module. After the projection, the visual tokens and textual tokens are denoted as $\boldsymbol{p}_{v} \in \mathbb{R}^{N_{v}\times  D} $ and $\boldsymbol{p}_{t} \in \mathbb{R}^{N_{t}\times  D} $, respectively. Then a learnable token (a learnable embedding) is attached, concatenated together with $\boldsymbol{p}_{v}$ and $\boldsymbol{p}_{t}$. The joint sequence is denoted as:
	\begin{equation}
		\boldsymbol{P} = [\ \underset{visual\ tokens \ \boldsymbol{p}_{v}}{\underbrace{\boldsymbol{p}_{v}^{1},\boldsymbol{p}_{v}^{2},\dots ,\boldsymbol{p}_{v}^{N_{v}}}}  , \underset{textual \ tokens \ \boldsymbol{p}_{t}}{\underbrace{\boldsymbol{p}_{t}^{1},\boldsymbol{p}_{t}^{2},\dots ,\boldsymbol{p}_{t}^{N_{t} }}}  ,\boldsymbol{p}_{l}], 
	\end{equation}
where $\boldsymbol{p}_{l} \in \mathbb{R}^{1\times  D}$ is the learnable token, which is randomly initialized before the training.
	
Next, a fusion transformer is used to embed $\boldsymbol{P}\in \mathbb{R}^{\left ( N_{v}+N_{t}+1 \right ) \times  D} $, which specifically includes 6 transformer encoder layers.

\textbf{Localization Module}. 
We use the representation of the learnable token from the multimodal fusion module as the input. The localization module is composed of a multi-layer perceptron (MLP), which is specifically composed of a 256-dim hidden layer, a ReLU activation function, and a linear layer, and outputs a 4-dimensional bounding box coordinate.

\subsection{Loss}
\label{sec:loss_fun}
Following the previous method \cite{Deng_2021_ICCV}, we apply the commonly used smooth L1 loss \cite{girshick2015fast} $\mathcal{L}_{smooth-L1} \left ( \cdot  \right ) $ and the generalized IoU (GIoU) loss \cite{rezatofighi2019generalized} $\mathcal{L}_{GIoU}\left ( \cdot  \right ) $ on the 4-dim bounding box coordinate. Adding GIoU loss is important for RSVG. The target size in RS images varies greatly, so the smooth L1 loss will be a large value when predicting a large box. Smooth L1 loss will be a small number when predicting a small box, even if the predicted box has a large error. Therefore, we normalize the coordinates of the ground-truth boxes according to the image size and use GIoU loss that is not affected by scale. Then, the whole loss function for training our proposed network can be written:
	\begin{equation}\label{eq:loss}
		\mathcal{L} = \mathcal{L}_{smooth-L_{1}}\left ( \boldsymbol{b}, \hat{\boldsymbol{b}}  \right ) + \lambda \cdot \mathcal{L}_{GIoU}\left ( \boldsymbol{b}, \hat{\boldsymbol{b}}  \right ), 
	\end{equation}
where $\boldsymbol{b}=\left ( x_{min}, y_{min}, x_{max}, y_{max} \right )$ denotes the coordinates of the ground-truth bounding box and $\hat{\boldsymbol{b}} =\left ( \hat{x} _{min}, \hat{y} _{min}, \hat{x} _{max}, \hat{y} _{max}  \right ) $ denotes the coordinates of the prediction. $\lambda$ is the hyper-parameter to balance two losses.

\section{Experiments}
\label{sec:experiments}
In this section, we present extensive experiments to validate the merits of our proposed MLCM. In Section \ref{sec:evaluation} and Section \ref{sec:implementation}, we introduce the evaluation metrics for RSVG and experimental setup details. We provide the main results of our method and compare the results with other state-of-the-art approaches for visual grounding in Section \ref{sec:results}. In Section \ref{sec:ablation}, we perform sufficient ablation experiments to verify the effectiveness of our MLCM. Finally, we show some qualitative results to fully analyze our model in Section \ref{sec:qualitative}.

\subsection{Evaluation metrics for RSVG}
\label{sec:evaluation}
Given an RS image-query pair, the predicted bounding box is considered right if the intersection-over-union (IoU) with the ground-truth bounding box is above a threshold. In previous visual grounding works, a threshold of 0.5 is used as an accuracy metric. We report the metrics with IoU thresholds at 0.5, 0.6, 0.7, 0.8, and 0.9, termed as Pr@0.5, Pr@0.6, Pr@0.7, Pr@0.8, and Pr@0.9, respectively. In addition, we follow the evaluation metrics of \cite{wu2020phrasecut}, including mean IoU and cumulative IoU (cumIoU), with the following equations:
	\begin{equation}\label{eq:mean_IoU}
		meanIoU = \frac{1}{M} {\textstyle \sum_{t}^{}}I_{t}    /U_{t},  
	\end{equation}
and	
	\begin{equation}\label{eq:cumulative_IoU}
		cumIoU = \left (  {\textstyle \sum_{t}^{}}I_{t}   \right ) /\left (  {\textstyle \sum_{t}^{}}U_{t}   \right ). 
	\end{equation}
Here $t$ is the index of image-query pairs and $M$ represents the size of the dataset. $I_{t}$ and $U_{t}$ are the intersection and union area between predicted and ground-truth bounding boxes.
	
\subsection{Implementation Details}
\label{sec:implementation}
We split the dataset by randomly assigning 40\%, 10\%, and 50\% of the expressions and their corresponding images to the training, validation, and test set. 
We resize the image size to a fixed size of 640$\times$640 for training. We set the maximum length of language tokens $N_{t}$ = 40 and the dimension $D$ = 256. 
The ResNet-50 and MLCM use the pre-training weights of the DETR model \cite{carion2020end}.
We use the pre-trained weights of BERT \cite{kenton2019bert} to initialize $BERT_{base}$ for textual feature extraction. The hidden size $b$ of BERT is 768. We follow the TransVG \cite{Deng_2021_ICCV} to process the input images and expressions. 
During training, we adopt AdamW \cite{loshchilov2018decoupled} with weight decay $10^{-4}$ as our optimizer. We train our network with a batch size of 8 for 150 epochs on one GTX 1080Ti 11GiB GPU. 
The dropout ratio is set to 0.1 for FFN in Transformer.
We set the initial learning rate of our network to $10^{-5}$ for pre-trained parameters and $10^{-4}$ for other parameters. We use Xavier \cite{glorot2010understanding} to randomly initialize the parameters without pre-training in our network.
For the loss function in Eq. \ref{eq:loss}, we set $\lambda$ = 1.

\subsection{Remote Sensing Image Visual Grounding Results}
\label{sec:results}
	
In order to assess the merits of our proposed method, we report our performance and compare it with the SOTA methods for natural image on our constructed RSVGD. As the results are shown in Table \ref{tab:results}, we observe that our method outperforms other works.
The two-stage method relies on a pre-trained object detector to generate object proposals and extract features, such as Mask R-CNN \cite{he2017mask}. Since the existing object detectors are pre-trained on natural images, the visual features of these detectors may not be compatible with the RSVG task. The quality of pre-generated proposals can be a performance bottleneck for the two-stage methods. 
The top parts of the table show results of current one-stage methods. The one-stage methods require pre-set anchors or manually designed complex multi-modal fusion mechanisms to yield bounding boxes. In fact, these works may lead to insufficient use of multi-modal information or over-fitting of datasets for specific scenes.
Our approach uses the transformer structure for feature encoding and feature fusion, which is more flexible and can realize more full interaction between visual information and textual information. Except for the Pr@0.9, our method is much higher than one-stage methods in other metrics. FAOA \cite{yang2019fast} fuses textual embeddings into YOLOv3 and fuses visual, textual, and spatial features at three different spatial resolutions. The model achieves the best accuracy at the threshold of 0.9 due to feature fusion at different resolutions, but the performance is still deficient at smaller thresholds. In the middle parts of Table \ref{tab:results}, we also compare our method to other transformer-based methods, \textit{i.e}, TransVG \cite{Deng_2021_ICCV} and VLTVG \cite{yang2022improving}. In contrast to our method, VLTVG designs a language-guided visual feature aggregation method and a multi-stage cross-modal decoder. The distinctiveness of visual features can be improved because visual features are concentrated in areas related to text descriptions while irrelevant areas are ignored in the training process. However, the performance is still insufficient because it ignores multi-level modality information. Our method follows the visual-linguistic transformer structure in TransVG to fuse multi-modal features. Besides, our MLCM uses multi-scale visual features and multi-granularity textual embeddings to learn more discriminative visual representations, which can aggregate effective information from multi-level multi-modal features and filter the redundant features of RS images.

	\begin{table*}[]
		\centering
		\caption{Comparison with the state-of-the-art methods for RSVG on the test set of RSVGD. The best performance is with bold and the second performance is with underline.}
		\begin{tabular}{ccccccccccc}
		\hline
		Methods         &Venue         & \begin{tabular}[c]{@{}c@{}}Visual\\ Encoder\end{tabular} & \begin{tabular}[c]{@{}c@{}}Language\\ Encoder\end{tabular} & Pr@0.5         & Pr@0.6         & Pr@0.7         & Pr@0.8         & Pr@0.9         & meanIoU        & cumIoU         \\ \hline
		\textit{\textbf{One-stage:}}       & &                &          &           &           &           &           &           &           &     \\
		ZSGNet \cite{sadhu2019zero}          & \textit{ICCV'19}& VGG                                                      & BiLSTM                                                     & 48.12          & 43.79          & 36.82          & 25.04          & 6.62           & 40.23          & 46.11          \\
		ZSGNet \cite{sadhu2019zero}         & \textit{ICCV'19}& ResNet-50                                                & BiLSTM                                                     & 51.67          & 48.13          & 42.30          & 32.41          & 10.15          & 44.12          & 51.65          \\
		FAOA-no Spatial \cite{yang2019fast} & \textit{ICCV'19}& DarkNet-53                                               & BERT                                                       & 63.63          & 61.20          & 56.92          & 50.15          & \textbf{38.83} & 57.53          & 62.66          \\
		FAOA \cite{yang2019fast}           & \textit{ICCV'19}& DarkNet-53                                               & BERT                                                       & 67.21          & 64.18          & 59.23          & 50.87          & 34.44          & 59.76          & 63.14          \\
		FAOA \cite{yang2019fast}           & \textit{ICCV'19}& DarkNet-53                                               & LSTM                                                       & 70.86          & 67.37          & 62.04          & 53.19          & \underline {36.44}    & 62.86          & 67.28          \\
		ReSC \cite{yang2020improving}           &\textit{ECCV'20} & DarkNet-53                                               & BERT                                                       & 72.71    & 68.92    & 63.01    & 53.70    & 33.37          & 64.24    & 68.10          \\

		LBYL-Net \cite{huang2021look}         & \textit{CVPR'21}& DarkNet-53                                                & LSTM                                                       & 73.29          &69.92           &  63.97         & 48.07          &  16.60         & 65.86          & 75.45    \\
			
		LBYL-Net \cite{huang2021look}         & \textit{CVPR'21}& DarkNet-53                                                & BERT                                                       & 73.78          & 69.22          & 65.56          & 47.89          & 15.69          & 65.92          &76.37     \\  \hline

		\textit{\textbf{Transformer-based:}}       & &                &          &           &           &           &           &           &           &     \\

		TransVG \cite{Deng_2021_ICCV}         & \textit{ICCV'21}& ResNet-50                                                & BERT                                                       & 72.41          & 67.38          & 60.05          & 49.10          & 27.84          & 63.56          & 76.27    \\
		
		VLTVG \cite{yang2022improving}         & \textit{CVPR'22}& ResNet-50                                                & BERT                                                       & 69.41          & 65.16          & 58.44          & 46.56          & 24.37          & 59.96          & 71.97    \\
		VLTVG \cite{yang2022improving}         & \textit{CVPR'22}& ResNet-101                                                & BERT                                                       & \underline{75.79}          & \underline{72.22}          & \underline{66.33}          & \underline{55.17}          & 33.11          & \underline{66.32}          & \underline{77.85}    \\

		\rowcolor[HTML]{EFEFEF}  
		Ours     & - & ResNet-50                                                & BERT                                                       & \textbf{76.78} & \textbf{72.68} & \textbf{66.74} & \textbf{56.42} & 35.07          & \textbf{68.04} & \textbf{78.41} \\ \hline
		\end{tabular} \label{tab:results}
	\end{table*}

\subsection{Ablation Study}
\label{sec:ablation}
In this section, we conduct detailed experiments to systematically analyze the proposed MLCM. As shown in Table \ref{tab:ablation}, we study the effectiveness of the multi-level cross-modal feature learning mechanism. The first row shows the RSVG results without multi-level cross-modal feature learning, which achieves 72.41\% Pr@0.5 on the testset of RSVGD. The second row shows the results of unimodal feature learning containing only multi-scale visual features, and the performance is dropped by 6.63\%. Then, we add sentence-level and word-level textual embeddings, respectively. The results, as shown in the third and fourth row, drop by 4.95\% and 1.51\%, respectively. In the fifth row, we adopt unimodal feature learning containing only multi-granularity textual embeddings, and the result is improved by 0.37\%. The last row indicates results of the complete multi-level cross-modal feature learning, showing a 4.37\% performance improvement over the absence of MLCM. The performance is greatly improved, which demonstrates that the representations for RSVG can be modeled more effectively with the design of MLCM module.

\begin{figure}[t]	
		\centering
		\includegraphics[width=0.85\linewidth]{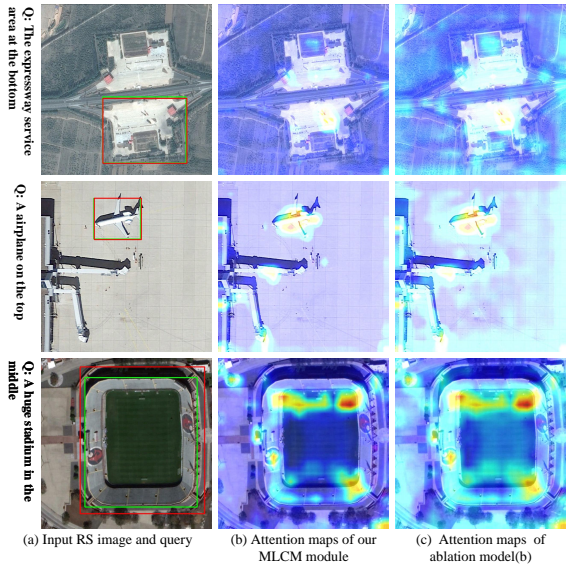}
		\caption{Visualization of the final grounding results (green/red boxes are ground-truths/predicted regions), the attention maps of our MLCM, and the attention maps of ablation model(b) for various input expressions and RS images on the RSVGD test set.
			}
		\label{fig:Ablation}
	\end{figure}

To deeply analyze the results containing only multi-scale visual features, we visually compare the attention map of MLCM and ablation model(b), as shown in Fig. \ref{fig:Ablation}. The darker background color in the attention map indicates the higher attention to this region. According to the attention maps, there are also many regions of dark color in the background or non-target areas when performing unimodal feature learning containing only multi-scale visual features. Therefore, due to the cluttered background of RS images, a large amount of noise is introduced when containing only multi-scale visual features. So the performance is greatly dropped. However, when multi-granularity textual embeddings are added, the noise in the attention maps is greatly filtered, so the performance is significantly improved. When only single-granularity textual embeddings (\textit{i.e.}, word-level or sentence-level) are added, some noise can be filtered, but the performance is still slightly lower than without MLCM. When unimodal feature learning containing only multi-granularity textual embeddings is performed, the performance is slightly improved than without MLCM. 
The above analyses prove that RS visual features are complex. But the MLCM module has multi-level multi-modal feature input and cross-modal learning capability to adaptively filter irrelevant noise, enhance salient features, and learn more discriminative visual representations.
	
	\begin{table}[]
		\centering
		\caption{The ablation studies of the MLCM module in our network.}
		\renewcommand\arraystretch{1.4}
		\begin{tabular}{c|c|cc|c}
			\hline
			\multirow{2}{*}{\begin{tabular}[c]{@{}c@{}}Models\end{tabular}}  & Visual      & \multicolumn{2}{c|}{Textual} & \multirow{2}{*}{\begin{tabular}[c]{@{}c@{}}Pr@0.5\\ (\%)\end{tabular}} \\
			& multi-scale & word-level  & sentence-level &                                                                        \\ \hline
			(a) & \XSolidBrush & \XSolidBrush &  \XSolidBrush  & 72.41                                                                  \\
			(b) & \CheckmarkBold   & \XSolidBrush    & \XSolidBrush      &  65.78$_{\downarrow 6.63 }  $                                                                \\
			(c) & \CheckmarkBold   &  \XSolidBrush    & \CheckmarkBold      &    67.46$_{\downarrow 4.95}         $                                                       \\
			(d) & \CheckmarkBold   & \CheckmarkBold   & \XSolidBrush  &   70.90$_{\downarrow 1.51}$                                                                    \\
			(e) & \XSolidBrush & \CheckmarkBold   & \CheckmarkBold      &   72.78$_{\uparrow 0.37}$                                                                  \\
			ours & \CheckmarkBold   & \CheckmarkBold   & \CheckmarkBold      & \textbf{76.78}$_{\uparrow 4.37}$                                                      \\ \hline
		\end{tabular} \label{tab:ablation}
	\end{table}

\subsection{Qualitative Results}
\label{sec:qualitative}
\begin{figure*}[tp]	
		\centering
		\includegraphics[width=0.85\linewidth]{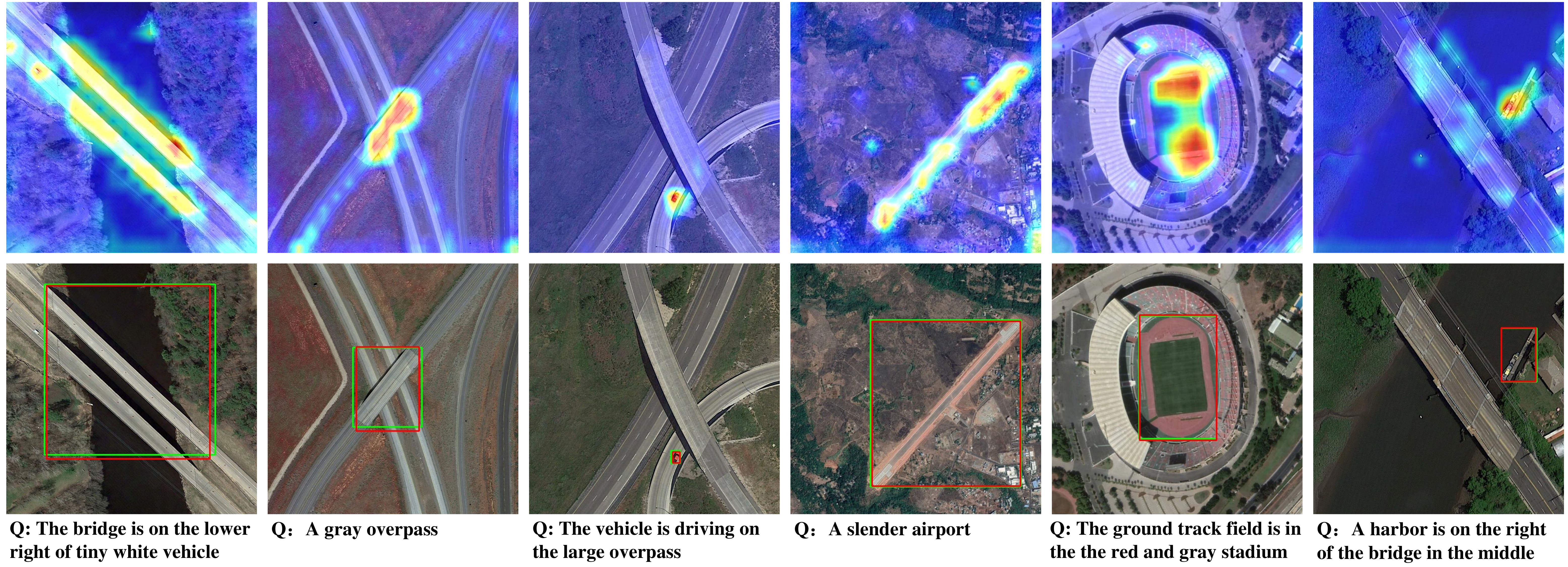}
		\caption{Visualization of the final grounding results and the attention maps of our proposed MLCM.
			}
		\label{fig:Qualitative_result}
	\end{figure*}

In Figs. \ref{fig:Ablation} and \ref{fig:Qualitative_result}, we show some qualitative results on the test set. We visualize the final grounding results and the attention maps for various inputs.
It is observed that our method can accurately localize objects described in query expressions with specific attributes. In addition, MLCM can focus on the visual features of the region where the target object is localized under the guidance of multi-level and multi-modal features. 
For example, the first three image-query pairs in Fig. \ref{fig:Qualitative_result} refer to a bridge with a vehicle, an overpass, and a vehicle driving on the overpass. MLCM accurately enhances the visual features of the corresponding areas of the bridge, overpass, and vehicle. MLCM can effectively generate interpretable attention of natural language corresponding to the shape and location of the entire target object.

\begin{figure*}[t]	
		\centering
		\includegraphics[width=0.85\linewidth]{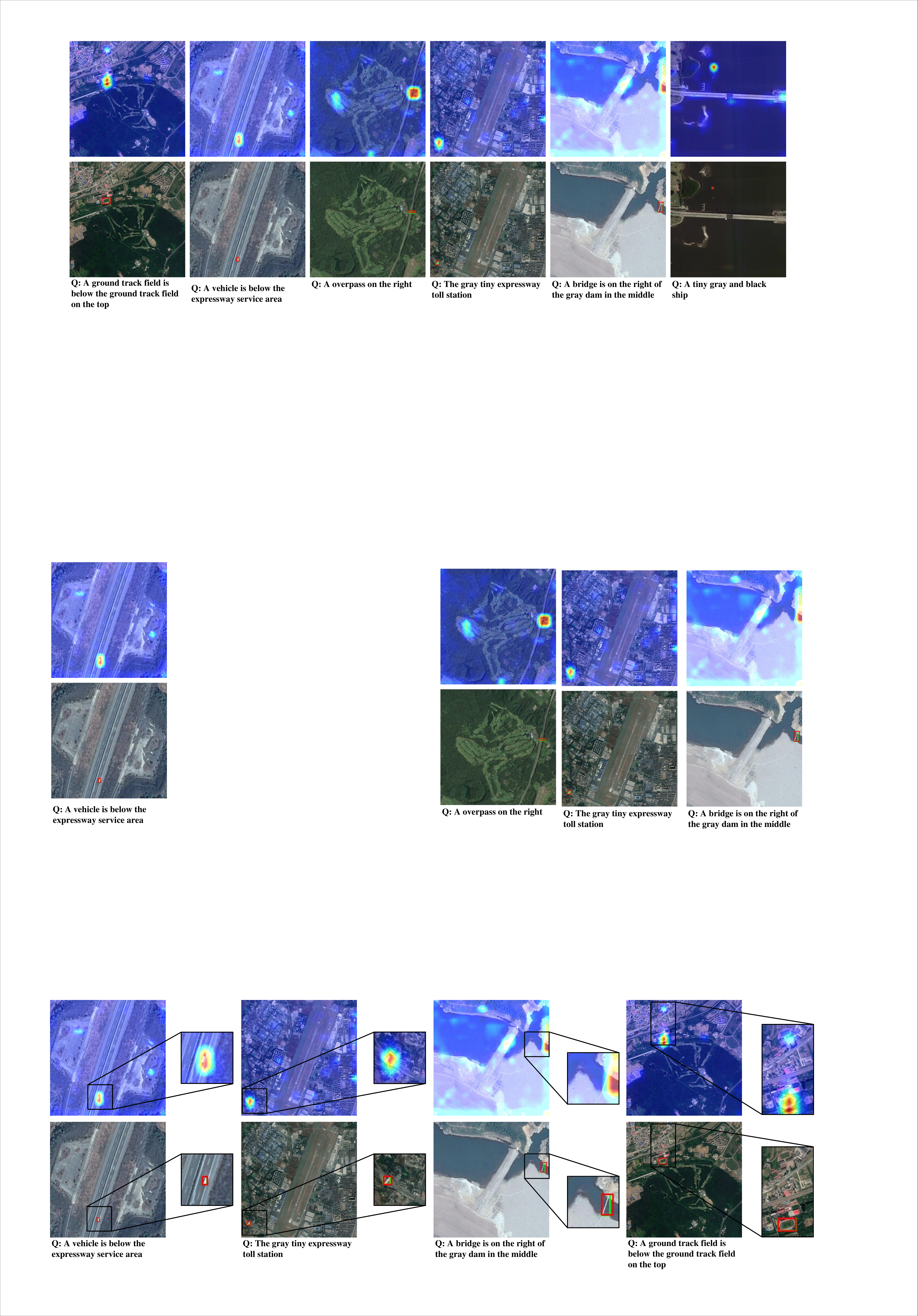}
		\caption{Some cases of small-scale target grounding on the RSVGD test set.
		}
		\label{fig:Small_scale}
	\end{figure*}

Due to the sufficient interaction of fine-granularity textual embeddings and multi-scale visual features, our method can accurately localize small-scale objects. The MLCM proposed in this paper has a better visual representation learning effect for small-scale targets. As shown in Fig. \ref{fig:Small_scale}, the dark backgrounds in the attention map are the regions where various small-scale objects are located. MLCM can combine multi-scale visual features and multi-granularity textual embeddings to precisely enhance the visual features of small-scale objects and improve the grounding accuracy.

\begin{figure}[t]	
		\centering
		\includegraphics[width=1\linewidth]{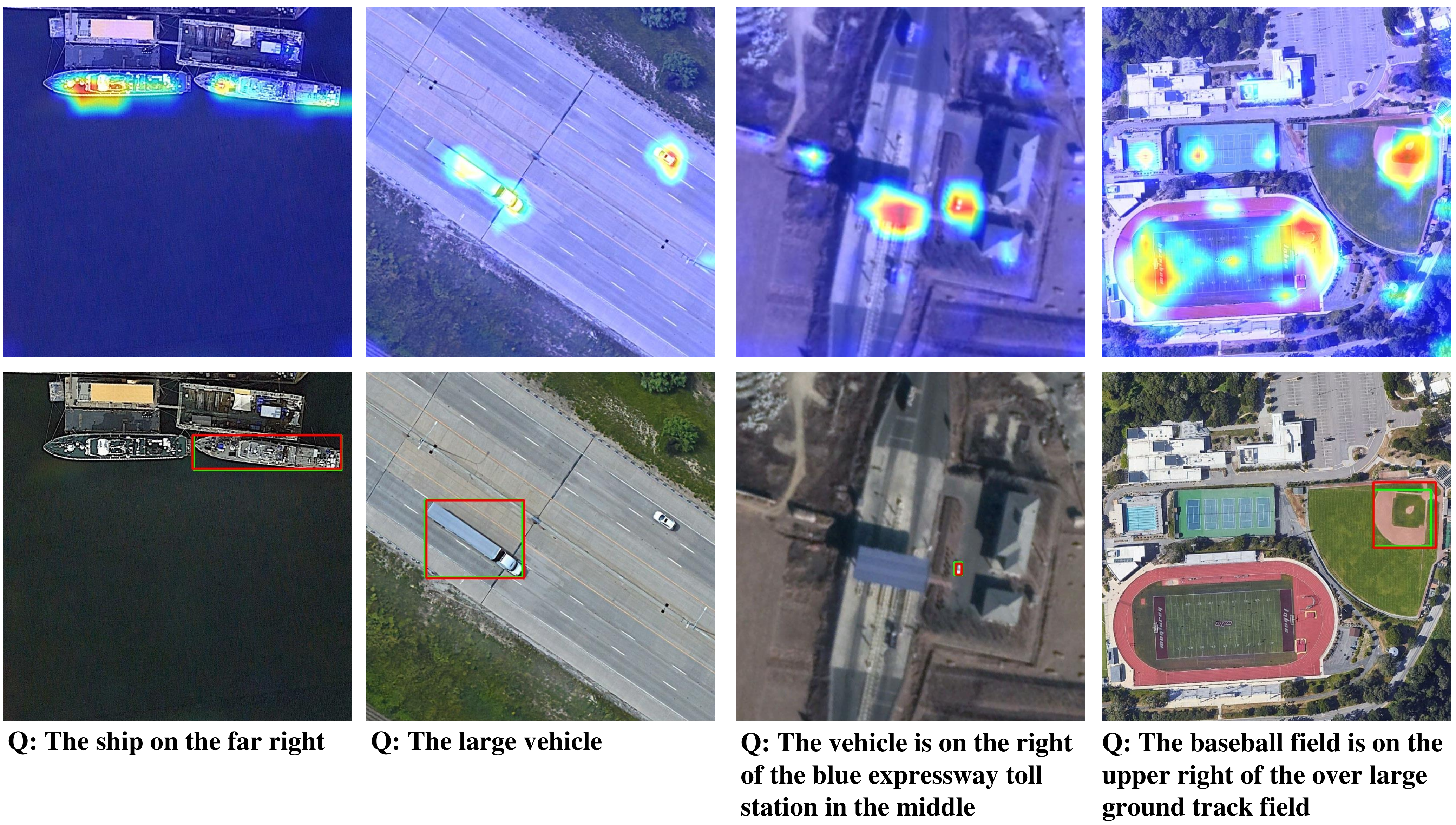}
		\caption{Some cases with wrong attention regions on the RSVGD test set.
		}
		\label{fig:enhanceError}
	\end{figure}

According to the visualization, many hot regions that are not the target regions are observed in attention maps. The regions of wrong attention are mainly divided into two types. The first type is the region where objects belonging to the same category as the target object are located, as shown in the first two data of Fig. \ref{fig:enhanceError}. The other is the region where objects that have a relationship with the target object are located. The second type exists in the last two data, which are the vehicle that has a relationship with an expressway toll station and the baseball field that has a relationship with a ground track field. However, the impact of wrong attention will be avoided in the transformer-based multimodal fusion module. The module makes full use of word-level textual embeddings with local semantic information to further align visual modality and textual modality for accurate visual grounding.

\begin{figure}[t]	
		\centering
		\includegraphics[width=0.8\linewidth]{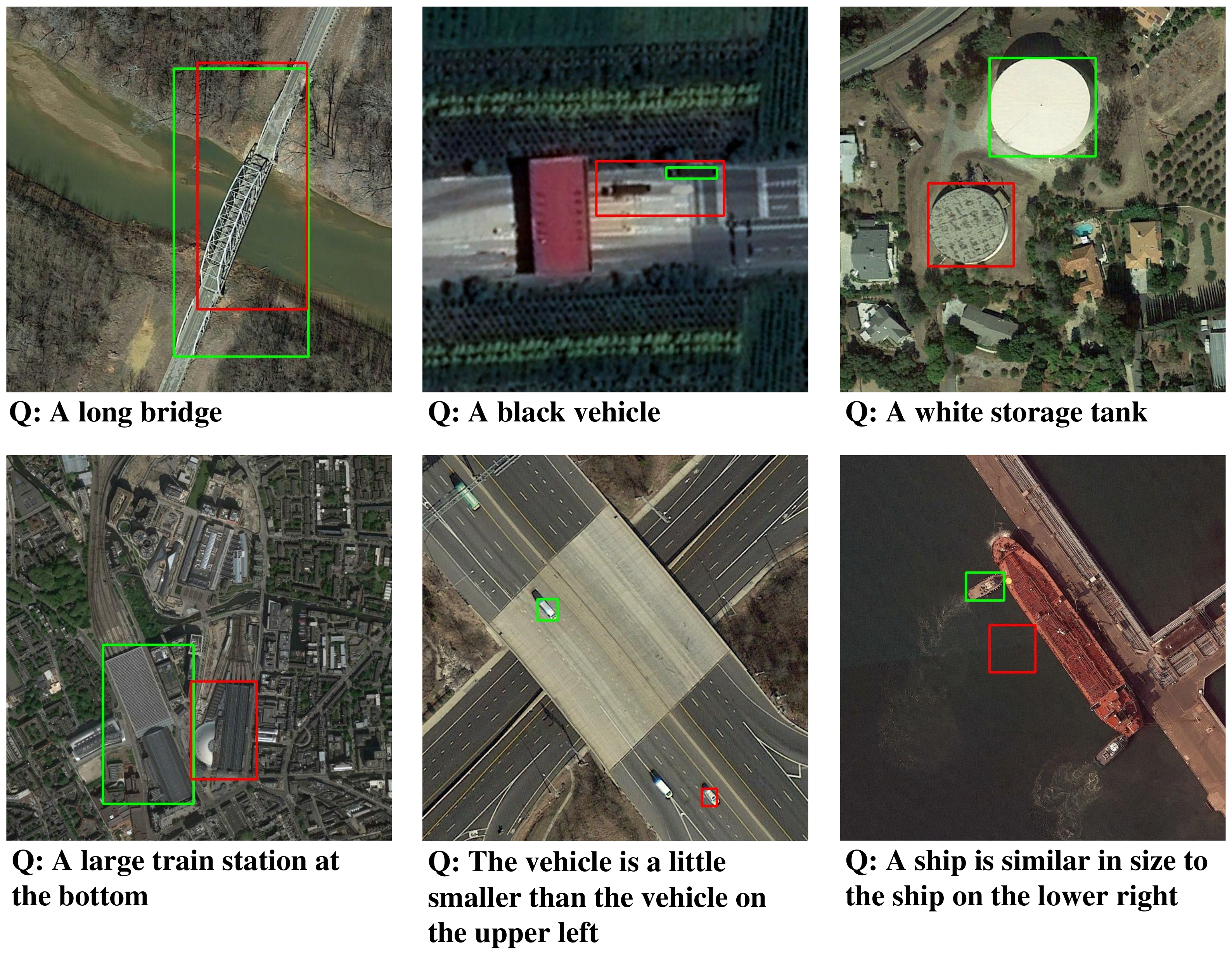}
		\caption{Failure cases of our method on the RSVGD test set.
			}
		\label{fig:error}
	\end{figure}

Our model is in line with human perception habits, which can learn more discriminative visual representations of RS images and effectively fuse and align visual features and textual embeddings. It can model and reason under the guidance of query expressions with complex relationships. However, our approach also has some failure results. As shown in Fig. \ref{fig:error}, there are three main types of grounding failure. 
The first type is shown in the first column. Due to the cluttered backgrounds of the RS image, there are areas or different objects with similar visual features to the target object and the scope of the object is difficult to define precisely.
The second is shown in the second column. Due to the incompleteness, complexity, and ambiguity of query expressions, the objects between the same category cannot be distinguished or the target objects cannot be clearly referred and localized.
The above problems in the dataset make it difficult for the model to accurately ground target objects, and errors will inevitably occur.
The last column of Fig. \ref{fig:error} is caused by the lack of performance of the model. When the object in the RS image is salient and the attributes described by the expression are clear and unambiguous, the model cannot complete the RSVG correctly. The result indicates that our proposed method still has some shortcomings and needs further research and improvement.

\section{Conclusion}
\label{sec:conclusion}
In this paper, we introduce a new task to ground natural language expressions on RS imagery. To the best of our knowledge, we build the new large-scale dataset for RSVG. RSVGD is obtained from the DIOR dataset by an automatic generation algorithm with manual verification, which greatly reduces the collection cost of the dataset and ensures the correctness of the dataset. RSVGD is large-scale on the number of image-query pairs and has high inter-class similarity and intra-class diversity. In addition, we benchmark extensive SOTA natural image methods on our constructed RSVGD and analyze the results. We obtain only acceptable results using natural image methods, suggesting the potential for future research.
Finally, a novel transformer-based MLCM module is devised to solve problems of the cluttered background and scale-variation of RS images. The main innovation is that MLCM adapts to multi-scale inputs and incorporates effective information from multi-level and multi-modal features to learn the attention of visual representations relevant to the query. Compared with existing natural image visual grounding methods, our approach achieves better performance and shows its superiority. In future work, more works need to be done on RSVG considering the characteristics of RS images.


\footnotesize
\bibliographystyle{IEEEtranN}  
\bibliography{document}  


%


\ifCLASSOPTIONcaptionsoff
  \newpage
\fi

\end{document}